# Driver Hand Localization and Grasp Analysis: A Vision-based Real-time Approach

Siddharth, Akshay Rangesh, Eshed Ohn-Bar, and Mohan M. Trivedi

*Abstract*—Extracting hand regions and their grasp information from images robustly in real-time is critical for occupants' safety and in-vehicular infotainment applications. It must however, be noted that naturalistic driving scenes suffer from rapidly changing illumination and occlusion. This is aggravated by the fact that hands are highly deformable objects, and change in appearance frequently. This work addresses the task of accurately localizing driver hands and classifying the grasp state of each hand. We use a fast ConvNet to first detect likely hand regions. Next, a pixel-based skin classifier that takes into account the global illumination changes is used to refine the hand detections and remove false positives. This step generates a pixel-level mask for each hand. Finally, we study each such masked regions and detect if the driver is grasping the wheel, or in some cases a mobile phone. Through evaluation we demonstrate that our method can outperform state-of-the-art pixel based hand detectors, while running faster (at 35 fps) than other deep ConvNet based frameworks even for grasp analysis. Hand mask cues are shown to be crucial when analyzing a set of driver hand gestures (wheel/mobile phone grasp and no-grasp) in naturalistic driving settings. The proposed detection and localization pipeline hence can act as a general framework for real-time hand detection and gesture classification.

*Index Terms*—In-cabin activity analysis, hand-object interaction, driver cell-phone usage, human-vehicle interaction, highly automated vehicles, control transition, takeover readiness.

## I. Introduction

Hand detection and gesture recognition are well-researched problems with wide range of applications in human-computer interaction and robotics. They have been employed for ego-centric applications [4, 5, 6], driver studies [3, 10, 13, 19, 26, 39, 42], and gesture based human-machine interaction [6, 13, 28] to name a few. Majority of the studies and their evaluation have been done only on datasets collected from same scene and camera viewpoint with minimal occlusion. In the context of intelligent vehicles, the changes in illumination are large and frequent. These changes occur rapidly depending on weather conditions, time of the day, and even where the vehicle is being driven (freeways/high rise cities). Fast and efficient detection of hands is desirable in such situations where safety of occupants is crucial. The work presented in this paper addresses these issues by presenting a fast hand detection and grasp recognition framework that is evaluated on real world driving data that presents us with all the challenges mentioned above. Specifically, we study state-of-the-art methods for hand detection, localization and object grasp recognition, and propose a system capable of working in real-time for a wide range of camera viewpoints, illumination changes, occlusions, and truncations.

Use of computer vision to study the environment inside and outside the vehicle is driven by many motivations. First, safety of the vehicle's occupants is directly correlated to driver's attentiveness. Engaging in distractions such as texting, interacting with gadgets inside vehicle and eating during driving is reportedly common [32]. Second, driver's hands provide a unique modality for understanding driver behavior [10, 26]. Driver behavior such as interaction with the steering wheel, gear shift or rear-view mirror provide subtle cues about the driver's experience and style. This style may change from driving on freeways versus driving in cities. The number of hands used to grasp the steering wheel and how firmly it is being grasped are additional cues that may indicate how comfortable the driver is during a given period. This makes hand detection and grasp recognition important aspects that contribute towards active safety. Third, in the context of passive safety, the driver's hands and reaction to pre-crash conditions provides insight into understanding the role of driver behavior in crashes.

To briefly outline the procedure: We first detect candidate hand regions in an image using YOLO [12] - a framework that enables real time object detection. The detections are then refined by passing them through a pixel-level skin classifier to segment the hand region. This saves us considerable computation since we do not search for skin regions in the whole image space. Finally, we provide grasp annotations (no grasp/grasping the wheel etc.) for each hand instance and train a classifier that predicts these labels at test time. To the best of our knowledge, this is fastest framework capable of doing robust hand detection, pixel-level localization and 3 class gesture recognition (wheel/mobile phone grasp or no grasp) in naturalistic driving scenarios.

## II. Related Research Studies

Hand detection has been researched and studied in the context of gesture analysis for a long time. Early efforts that were used to detect human hands from colored images mostly concentrated on skin based features (either color, texture or both) [23, 24, 25]. Color of the skin was used as the major feature for detecting hands. Skin color was modeled by various techniques such as mixture of Gaussians [33] or multi-layer perceptron classifier. However, these approaches fail in dealing with a wide variety of changes in illumination or in working on unfamiliar scenes on which

Siddharth is with Department of Electrical and Computer Engineering at University of California, San Diego, La Jolla, CA 92093 USA (e-mail: sid.sapien@gmail.com).
Akshay Rangesh, Eshed Ohn-Bar, and Mohan M. Trivedi are with the Laboratory for Intelligent and Safe Automobiles (LISA), University of California San Diego, La Jolla, CA 92093-0434 USA (email: arangesh@ucsd.edu; eohnbar@ucsd.edu; mtrivedi@ucsd.edu).

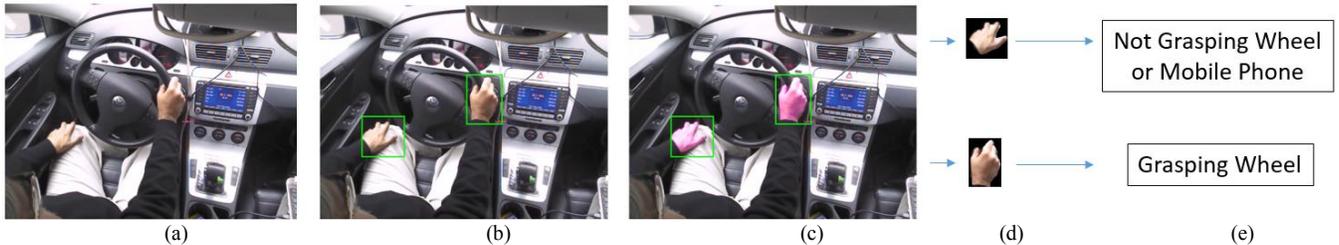

Figure 1. An overview of the proposed framework: (a) Original image (b) Detection of hands (c) Pixel-level skin classification (d) Localized hand extraction (e) Object grasp/No-grasp classification

they have not been trained beforehand. The wide range of changes in illumination are frequent in the context of naturalistic driving settings.

Also, these approaches generally require the pixel-level classification of skin and non-skin regions on the whole image, which is highly inefficient considering only a small part of an image actually comprises of hand(s).

Recent works such as [1, 5] and specifically that by Li and Kitani [4] have improved on such pixel-level classifiers by modeling the global appearance of hands and using t-SNE [29] to visualize its dependence on color and texture based features. These global appearance models learn the changes in appearance due to varying illumination. However, this method has only been tested on egocentric videos [1, 4, 5, 6] where the illumination does not vary as much as it does in unconstrained outdoor environments.

The advances in object detection and recognition have inspired researchers to employ local appearance based features that can be extracted from images to train Viola and Jones-like boosted detector [21], HOG-SVM detector and classifier [7, 9] or Fast Feature Pyramids [2]. In our experience, such detectors do not work very well for the wide range of hand poses that are common in real-life. The issue might be their inability to learn such a wide variety of deformations in hand templates. For hand detection in videos, motion-based methods are generally employed with the assumption that hands (foreground) and the background have different motion or appearance statistics. This cannot be assumed in case of driver's hands due to lack of continuous motion while grasping the wheel and hence the hands may end up being classified as background.

The recent advances in deep learning and Convolutional Neural Networks (ConvNets) have made it possible to train deep neural networks for object detection, [12, 35] and particularly hand detection [1, 34]. These methods generally perform much better than appearance (either local or global) and motion based approaches. This may be attributed to the fact that ConvNets are highly non-linear models, capable of learning "high-level" features [40, 41] in addition to "low-level" skin based features. By employing such "high-level" features these methods can even deal with occlusion.

Very recently, the You Only Look Once (YOLO) [12] object detection framework based on ConvNets has caught the attention of research community, mostly due to its real-time performance. It outperforms other detection frameworks in speed and holds excellent generalization capabilities from natural images to other domains. Although it is known to suffer more from localization errors than other such detectors, is less likely to predict false positives.

Hand gesture recognition [6, 22] requires proper localization of hands. If localization of hands is accurate, then gradient based methods such as HOG [7] can yield good results [6] for clustering the hand gestures. Frequency based methods have also been used on localized hand regions for gesture recognition [20]. In the context of driver's hands, knowing that the driver is grasping the wheel or not, and using which hand to grasp can be of utmost importance [10, 26]. It can be used to predict and interpret the driver's behavior from motion patterns [19].

**Achieving Fast and Robust Performance in Naturalistic Driving Settings:** This work proposes a combination of the pixel based hand detector [4] and YOLO in order to improve hand localization performance. We first study the state-of-the-art pixel based hand detector [4] in context of in-vehicle driving images. As the method is shown to perform poorly, the YOLO framework for hand detection is used to achieve better robustness in driving videos with wide variations in illumination, occlusion, scene and camera position. Nonetheless, as YOLO suffers from localization errors, the pixel-based skin detector is employed with both local and global appearance based features in order to refine the detections from YOLO framework. This increases true positives (by increasing the overlap between detections and ground truth) and removes false positives (by removing detections having no skin area). Additionally, incorporating different global illumination models provides further improvement in hand localization performance. The resulting detector is fast, accurately localizes the hands, and provides a hand mask suitable for extracting grasp cues. The refined hand masks are consequently used for determining driver's wheel or mobile phone grasping activity.

### III. PROPOSED METHOD

Our goal is to discover hand instances and perform object grasp/no-grasp classification on in-vehicular images captured from different viewpoints. This is accomplished in five steps: (1) Extracting local appearance based features (2) Global appearance modeling to represent changes in illumination (3) ConvNet based detection pipeline (4) Pixel-level hand localization (5) Extracting features from localized hand of the driver for object grasp/no-grasp classification.

#### A. Extracting Local Appearance Based Features

As mentioned above in Section II, color and texture based features are very important in distinguishing between skin and non-skin regions. We take cues from previous work [4, 23, 24, and 25] in pixel-based hand detection which shows that RGB, LAB and HSV color-spaces show robust performance in detection of skin regions. Like the works form Kitani [4] and Wong [5], we also experimented on our dataset with the use of other color and texture based features such as HOG [7], BRIEF [15], ORB [16], SIFT [17], SURF [18], and superpixels [36]. For real-time performance we chose to go forward with RGB, HSV and SIFT based features only as they proved to be calculated faster with good performance. Specifically, we looked into how local color information from HSV and LAB color-spaces contribute towards detection performance. We used the 128 dimensional SIFT descriptor based on gradient histograms to capture local contours for classification (skin or non-skin) performance.

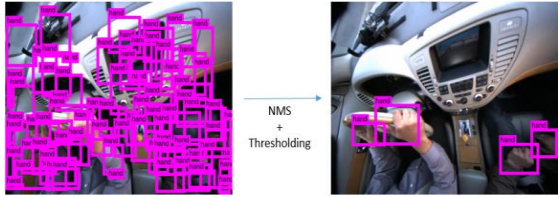

Figure 2. Detections from YOLO are filtered using NMS and confidence threshold.

### B. Global Appearance Modeling

Like [4] we modeled the global appearance of hands to represent the changes in illumination. This is very important since during naturalistic driving, the illumination changes rapidly and with high variations. We group 10 illumination models indexed by a global color histogram to learn their global illumination by k-means clustering on the HSV histograms of each training image. A separate random tree regressor [30] for each cluster is learned. The underlying assumption is that hands with similar global appearance will share a similar distribution in feature space and will be clustered together.

### C. Generating Hand Detection Proposals

We use the YOLO [12] based deep convolutional neural network for detection of hands in images. YOLO is faster compared to other such ConvNet based frameworks such as FRCNN [37] and Adaptive Region Pooling [38]. We used the weights pre-trained on Pascal VOC Dataset [11] for training on our dataset (Section IV). We use 24 convolutional layers followed by 2 fully connected layers for training YOLO with learning rate of 0.0005, momentum of 0.9, and dropout rate of 0.5. We choose detections above a certain confidence threshold for further processing, so as to save computation time by not searching for skin regions in all the detections. In Fig. 2 we show detection proposals from YOLO which we then filter using Non Maximum Suppression (NMS) and thresholding based on confidence scores to get the final detections.

### D. Refining Detection Proposals

As discussed before in Section II, YOLO is very fast but suffers from poor localization. We use local and global appearance based features discussed in Sections IIIB and IIIC above for fine-grained classification of skin/non-skin regions inside the detection boxes. This helps to remove some false positives when no skin part is detected inside the detection. Fig. 3 shows some detections by YOLO followed by the pixel-level hand masks.

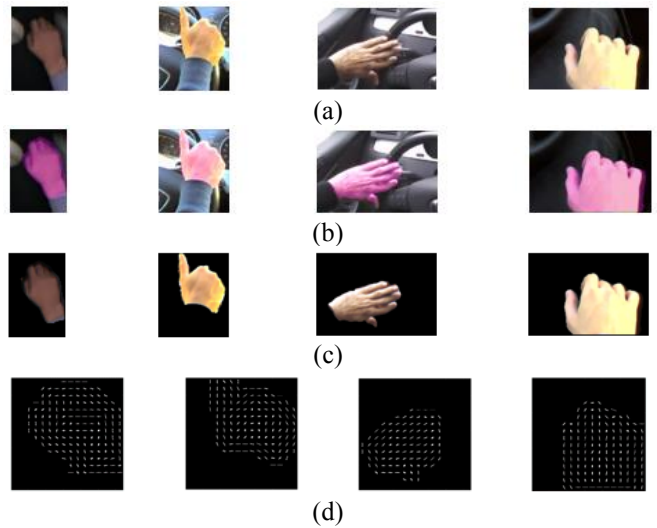

Figure 3. (a) Detections by YOLO. (b) Pixel based skin detector. (c) Masked hand descriptor. (d) Masked hand HOG descriptor (resized).

### E. Object Grasp/No-Grasp Classification

For the images of driver's hands, we obtain hand masks by the above procedure and use the localized hand regions after masking them. Masking enables us to remove all the nearby background from the detection before extracting features. We then resize all the detections to 128x128 pixels which we found as a good estimate of size based on the driver's hand instances in the dataset.

We use HOG descriptor for the masked images in Fig. 3(c) to extract oriented gradient features from the hand regions (shown in Fig. 3(d)). We use the following HOG template parameters: 8x8 cell, 8x8 stride, 16x16 blocks with 9 gradient orientation bins. The large block size is used to reduce the ability to suppress local illumination changes since we have already taken care of them for hand region detection before. The HOG feature vector is reduced to 30 taking only the most informative features using PCA [27]. We then use an SVM [31] based classifier trained for a two-class problem, dividing dataset into 80-20 ratio for training and testing. We use the linear kernel for training the SVM classifier with ten-fold cross validation. Below (section IV.C), we show two types of HOG-SVM evaluations. One involves extracting HOG features directly from detections (Fig. 3(a)), and other using HOG features from the hand mask (Fig. 3(d)).

## IV. EXPERIMENTAL EVALUATION

We first evaluate the use of pixel-based hand detector on the whole in-vehicular image using local and global appearance features. The method is slow since it has to search every pixel in the image. We then evaluate the YOLO based hand detection and employ pixel-based detector only on the detections by YOLO. Hence, the search space is reduced by a factor of 15 to 30 in the image depending on the size of detected hands. We then follow the pipeline for localizing and grasp/no-grasp classification as detailed above. Finally, we evaluate the performance of object grasp/no-grasp classification for driver's hands.

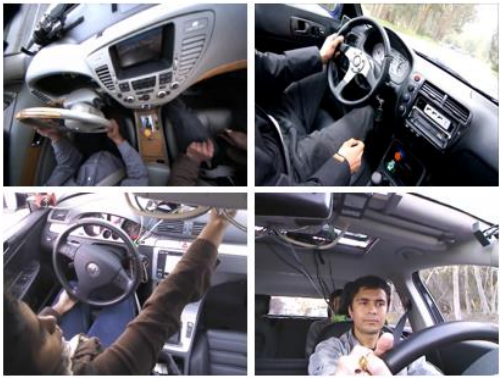

(a)

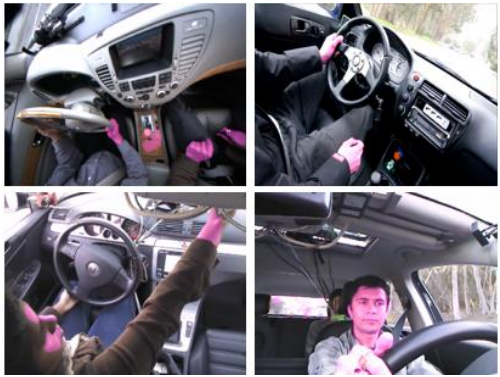

(b)

Figure 4. (a) Some in-vehicle images from the dataset. (b) Using only pixel based detector on the whole image.

*A. Dataset*

We used VIVA Hands Detection Dataset [10] for evaluation which has naturalistic driving images captured using many different camera viewpoints. The images have been captured using multiple vehicles, during varying illumination, hand occlusion and truncation, and also taken from YouTube. The hand instances have been divided into two parts- L1 and L2 for evaluation. L1 hand instances have a minimum height of 70 pixels and are from only over the shoulder (back) camera view. The L2 hand instances are smaller with minimum height of 25 pixels and consists of hand instances from all camera views. Fig. 4(a) shows some of the hand instances in the dataset. The dataset has 5500 images for training and 5500 images for testing both containing images with varying image dimensions.

In addition to testing the detector on VIVA dataset we also use the EDSH Dataset [4, 6] for training pixel-based skin classifier. The dataset has over 200 million labeled pixels for hand and non-hand regions. Furthermore, we labeled hand and non-hand regions in 550 images from training set of VIVA dataset using Grabcut [8]. Hence, in total we have more than 500 million labeled hand and non-hand pixels. We then label 1048 randomly chosen driver's hand instances (727 for grasping wheel, 110 for grasping mobile phone and

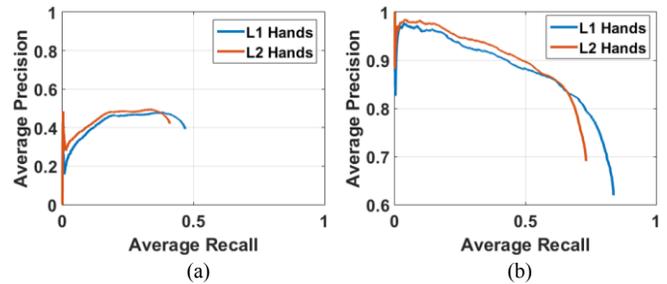

(a) (b)

Figure 5. (a) Precision-Recall curve for pixel-based hand detector (b) Precision-Recall curve for refined YOLO detections.

211 for grasping no object) from true labels in the training set of VIVA dataset and jitter them by scaling and translation to form a dataset of 5240 hand instances which we divide in an 80:20 split for training and testing to classify as object (wheel or mobile phone) grasp/no-grasp. We use a 6 cores, 3.5 GHz machine with 16GB RAM and Titan X GPU for training and testing of our framework particularly for the training and testing the deep neural network framework.

*B. Hand Detection Evaluation*

We first evaluate the pixel based hand detector. In Fig. 4(b) we show some of the hand instances labeled by pixel based detector. As we can see from Fig. 4(b), the method fails when the color and texture of in-vehicular parts such as music system, visor, driver's clothes etc. match that of skin. These provide many false detections. Also, when the illumination is very high on any of the hands, then the pixel based detector fails again to detect them. Furthermore, since pixel based detector only detects skin, the face and forearm are also output as true detections. Fig. 5(a) shows the precision-recall curve for pixel based detector evaluation. It is clear from the curve that the pixel based detector fails miserably due to above mentioned reasons. We believe that such a detector would perform better in egocentric camera based videos where the only skin-region visible in front of camera are hands and both hands have similar illumination characteristics at any given time. The average precision and recall values for L1 hand instances were 19.7 and 3.6 while for L2 hand instances were 18.2 and 2.8.

We then evaluate YOLO based detections and use pixel-based detector on detections above a confidence threshold (0.15) for increasing the overlap between hand's true bounding box and detection. This helps in surpassing the 0.5 overlap criterion set for the detections on test dataset. The method is robust as can been seen from precision-recall curves in Fig. 5(b). Additionally, this framework is fast and performs at up to 35 frames per second. The average precision and recall values using this method for L1 hand instances were 74.1 and 47.2 while for L2 hand instances were 66.9 and 40.2.

*C. Object (Wheel/Mobile Phone) Grasp Classification*

We show some of the driver's hand instances for wheel grasping, mobile phone grasping and no grasp activity in Fig. 6. As we notice from Fig.6, drivers may tend to have soft grasp by just keeping hand on the wheel or may fully grasp it. The other hand of the driver too may rest in a grasping gesture or may grasp a mobile phone. The camera viewpoints are too different for grasping mobile phone instances. These factors make the grasp classification problem hard. We first evaluate the grasp classification problem as a 2 class one where positive class is grasping the wheel and negative class is not grasping the wheel. First, we extract HOG based features from the detections as shown in Fig. 3(a) above and then we perform pixel-level classification on hand instances to extract HOG features from masked hand images as in Fig. 3(e). We then use off-the shelf VGG based features [40, 41] since they are handy for extracting useful information from the images. Similar to HOG, we extract VGG features before and after masking the hand instances.

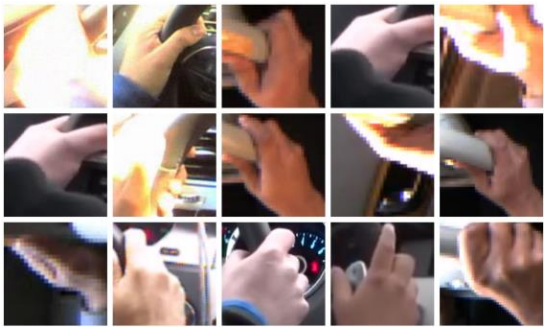
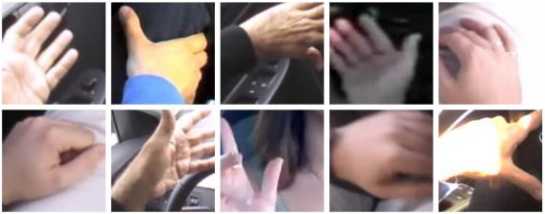
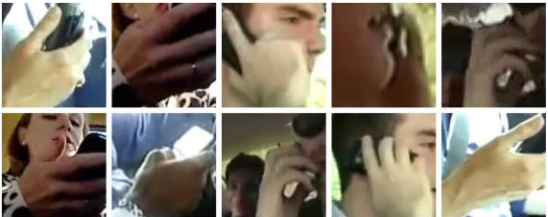

Figure 6. Examples of Driver's jittered hand instances used for training grasp classifier (a) Wheel Grasp instances. (b) Non-Wheel Grasp instances. (c) Mobile Phone Grasp instances.

Table I. Table showing performance comparison for 2 classes (Grasping wheel or not grasping wheel) and 3 classes (Grasping wheel, grasping mobile phone, and not grasping either) using HOG and VGG features before and after masking hand instances.

| Classification Performance for 2-Classes | | | |
|---|---|---|---|
| *Features Used* | *Wheel Grasp Accuracy* | *No Wheel Grasp Accuracy* | *Overall Accuracy* |
| **Detection + HOG** | 0.77 | 0.21 | 0.70 |
| **Hand Masking + HOG** | 0.80 | 0.69 | 0.795 |
| **Detection + VGG** | 0.78 | 0.28 | 0.66 |
| **Hand Masking + VGG** | 0.74 | 0.05 | 0.67 |
| **Classification Performance for 3-Classes** | | | |
| *Features Used* | *Wheel Grasp Accuracy* | *Mobile Phone Grasp Accuracy* | *No Object Grasp Accuracy* | *Overall Accuracy* |
| **Detection + HOG** | 0.67 | 0.04 | 0.15 | 0.53 |
| **Hand Masking + HOG** | 0.73 | 0.19 | 0.18 | 0.65 |
| **Detection + VGG** | 0.79 | 0.29 | 0.13 | 0.66 |
| **Hand Masking + VGG** | 0.79 | 0.53 | 0.38 | 0.73 |

We then apply PCA to only get 30 most informative HOG and VGG features for training SVM as detailed in Section III.E. As we can see from first part of Table I, for both HOG and VGG features, the classification performance is better with features extracted from hand mask images, compared to those extracted directly from detections. HOG features perform much better (79.5 accuracy) than VGG features (67% accuracy) for 2-class problem.

Additionally, we extend the two class problem to three classes, making a different class for mobile phone grasping hand instances. We extract the features similarly before and after hand masking while training a one vs all SVM framework for classification. We again find that for both HOG and VGG features, hand mask estimation increases performance for all the three classes. This further validates the use of refined hand masks obtained from the pixel-based detector for hand gesture classification. For the three class problem, VGG features, obtained from refined hand instances (73% accuracy) outperforms HOG based features computed on the same hand instances (65%).

We also see from Table I that how the false positives are reduced considerably for all the classes after refinement. Grasp classification, being a vital cue that can improve occupants' safety, the increased performance may prove crucial in saving lives.

## V. CONCLUSION

In this paper, we have studied the viability of using a fast ConvNet based framework for hand detection in naturalistic driving videos, coupled with s pixel-based hand detector for accurately localizing hand instances. A novel, naturalistic dataset for grasp analysis was introduced. Furthermore, a framework for grasp classification using pixel-based classifier was proposed. Additionally, the proposal runs in real-time (~35fps) which gives it a big advantage over state-of-the-art hand activity detectors and gesture classifiers. Finally, the framework can be employed for applications in hand detection, localization and gesture recognition for non-driving scenes too.

## VI. Acknowledgments

The authors would like to acknowledge the support of our sponsors and associated industry partners, and our UCSD LISA colleagues who helped in a variety of important ways in our research studies.

## VI. Acknowledgments

The authors would like to acknowledge the support of our sponsors and associated industry partners, and our UCSD LISA colleagues who helped in a variety of important ways in our research studies.